\def\year{2022}\relax
\documentclass[letterpaper]{article} 
\usepackage{aaai22}  
\usepackage{times}  
\usepackage{helvet}  
\usepackage{courier}  
\usepackage[hyphens]{url}  
\usepackage{graphicx} 
\usepackage{natbib}  
\usepackage{caption} 
\DeclareCaptionStyle{ruled}{labelfont=normalfont,labelsep=colon,strut=off} 
\usepackage{algorithm}
\usepackage{algorithmic}

\usepackage{newfloat}
\usepackage{listings}
\floatstyle{ruled}
\newfloat{listing}{tb}{lst}{}
\floatname{listing}{Listing}
\usepackage{kotex}
\usepackage[table,xcdraw]{xcolor}
\usepackage{multirow}
\usepackage{graphicx}
\usepackage{bm}
\usepackage{amsmath}
\usepackage{amssymb}

\usepackage{multicol}
\usepackage{caption}
\usepackage{subcaption}

\title{Multimodal Interactions Using Pretrained Unimodal Models for SIMMC 2.0}
\author {
    Joosung Lee, Kijong Han
}
\affiliations{
    \{rung.joo, mat.h\}@kakaoenterprise.com
}

\usepackage{bibentry}

\begin{document}
\maketitle

\begin{abstract}
This paper presents our work on the Situated Interactive MultiModal Conversations 2.0 challenge held at Dialog State Tracking Challenge 10. SIMMC 2.0 includes 4 subtasks, and we introduce our multimodal approaches for the subtask \#1, \#2 and the generation of subtask \#4. SIMMC 2.0 dataset is a multimodal dataset containing image and text information, which is more challenging than the problem of only text-based conversations because it must be solved by understanding the relationship between image and text. Therefore, since there is a limit to solving only text models such as BERT or GPT2, we propose a multimodal model combining image and text. We first pretrain the multimodal model to understand the relationship between image and text, then finetune our model for each task. We achieve the 3rd best performance in subtask \#1, \#2 and a runner-up in the generation of subtask \#4. The source code is available at https://github.com/rungjoo/simmc2.0.

\end{abstract}

\section{Introduction}
SIMMC 1.0 challenge~\cite{moon-etal-2020-situated} held at DSTC9 aims to build a real word assistant that handles multimodal inputs. SIMMC 2.0 dataset~\cite{kottur2021simmc}, which is an extension of SIMMC 1.0, is a dataset that is closer to the real-world context in a fashion or furniture shopping scenario, opened in DSTC10. SIMMC 2.0 dataset consists of task-oriented dialogs based on photo-realistic VR scenes in the shopping domain.

In addition to the SIMMC challenge, multimodal data sets based on image and text information are of great interest in the AI community. VisDial~\cite{Das_2017_CVPR} is a multimodal task that responds appropriately when given a visual-dialog image, conversation history, and a question. ~\citet{Vries_2017_CVPR} introduces a multimodal task that finds objects for sequences of queries in a rich image scene. ~\citet{DeepFashion2} introduces a comprehensive fashion data set containing categories of clothing in commercial shopping stores, providing a description of the object.

Multimodal models for various tasks with both image and text input have been widely studied. ~\citet{huang2021joint} and ~\citet{tom} achieve good performance in SIMMC 1.0 by using visual information as a text description of the object. However, these approaches have limitations in that object descriptions must be given, and it is difficult to regard them as a multimodal model using image information. MTN~\cite{le-etal-2019-multimodal} achieved the best performance in DSTC 7 (Video Scene-aware Dialogue)~\cite{yoshino2019dialog} and is a baseline model of SIMMC 2.0. MTN extracts each feature for each modal and generates an appropriate response through a transformer~\cite{NIPS2017_3f5ee243}. ~\citet{MurahariBPD20} improves performance by pretraining VilBERT~\cite{NEURIPS2019_c74d97b0} for Conceptual Captions and Visual Question Answering and finetuning for VisDial. CLIP~\cite{CLIP} is trained to match the transformer between image and text description, so it has the advantage of text-image matching for the unseen object class.

SIMMC 2.0 challenge requires understanding the relationship between image and text information and consists of the following 4 subtasks for multimodal conversational reasoning: 1) Multimodal Disambiguation 2) Multimodal Coreference Resolution 3) Multimodal Dialog State Tracking 4) Multimodal Dialog Response Generation \& Retrieval. Details are introduced in Section~\ref{sec:task_description}. We introduce an approach to subtask \#1, \#2, and subtask \#4-generation. We observed some of the data that could not be predicted by the unimodal model alone in all subtasks. Therefore, we introduce a multimodal model that combines pretrained unimodal models, which are first pretrained to understand the image and text representations of each other. In the pretraining stage, two models are trained: 1) a model that matches the image and text description of the cropped object, and 2) a model that matches the background of the object and context. We then use the representations of these two models to finetune the multimodal model for each subtask.

\section{Task and Data Description}
\label{sec:task_description}
The challenge consists of 4 subtasks, and the information of test data available for each subtask is different. When testing all tasks, the visual metadata of the object, the annotated information of the user, and the list of all objects referenced in the dialog (\textit{mention\_inform}) are unavailable. The object's nonvisual metadata, the objects referenced by the system utterance, the background image corresponding to the dialog, and the bounding box of all objects are available. 1) \textbf{Multimodal disambiguation} is a task of predicting multimodal disambiguation given the conversation history and the current user's utterances. It is a binary classification problem as true or false, and the object corresponding to the user's utterance is unknown. 2) \textbf{Multimodal Coreference Resolution} is a task of matching objects referenced in the user's utterance. During testing, the same data as subtask \#1 is available. 3) \textbf{Multimodal Dialog State Tracking} is a task to track dialog states based on multimodal context, which has a similar goal to traditional text-based dialog state tracking but also considers image input. During testing, the same data as subtask \#1 is available. 4) \textbf{Multimodal Dialog Response Generation \& Retrieval} is the task of generating or retrieving an appropriate utterance for the user. Subtask \#4 assumes that it has actual meta information (\textit{system\_transcript\_annotated}), so the objects referenced by the current turn's system are available. The data used for our model in the training and testing phases are described in Section~\ref{sec:data_ours}.

\section{Method}

\begin{figure*}
    \centering
    \begin{subfigure}{.5\textwidth}
      \centering
      \includegraphics[width=0.9\linewidth]{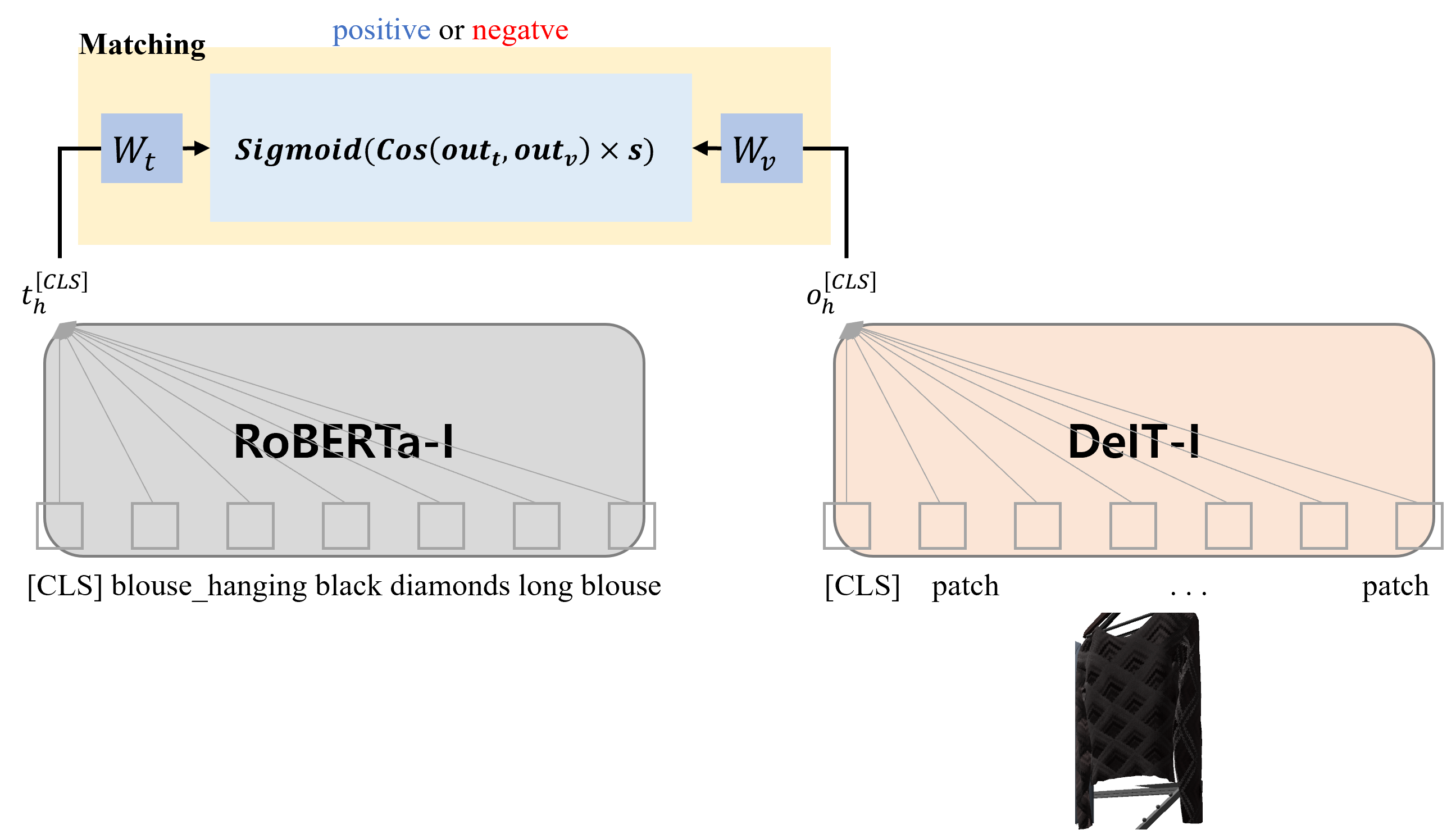}
      \caption{The structure of Image-to-text matching (ITM).}
      \label{fig:sfig1}
    \end{subfigure}%
    \begin{subfigure}{.5\textwidth}
      \centering
      \includegraphics[width=0.9\linewidth]{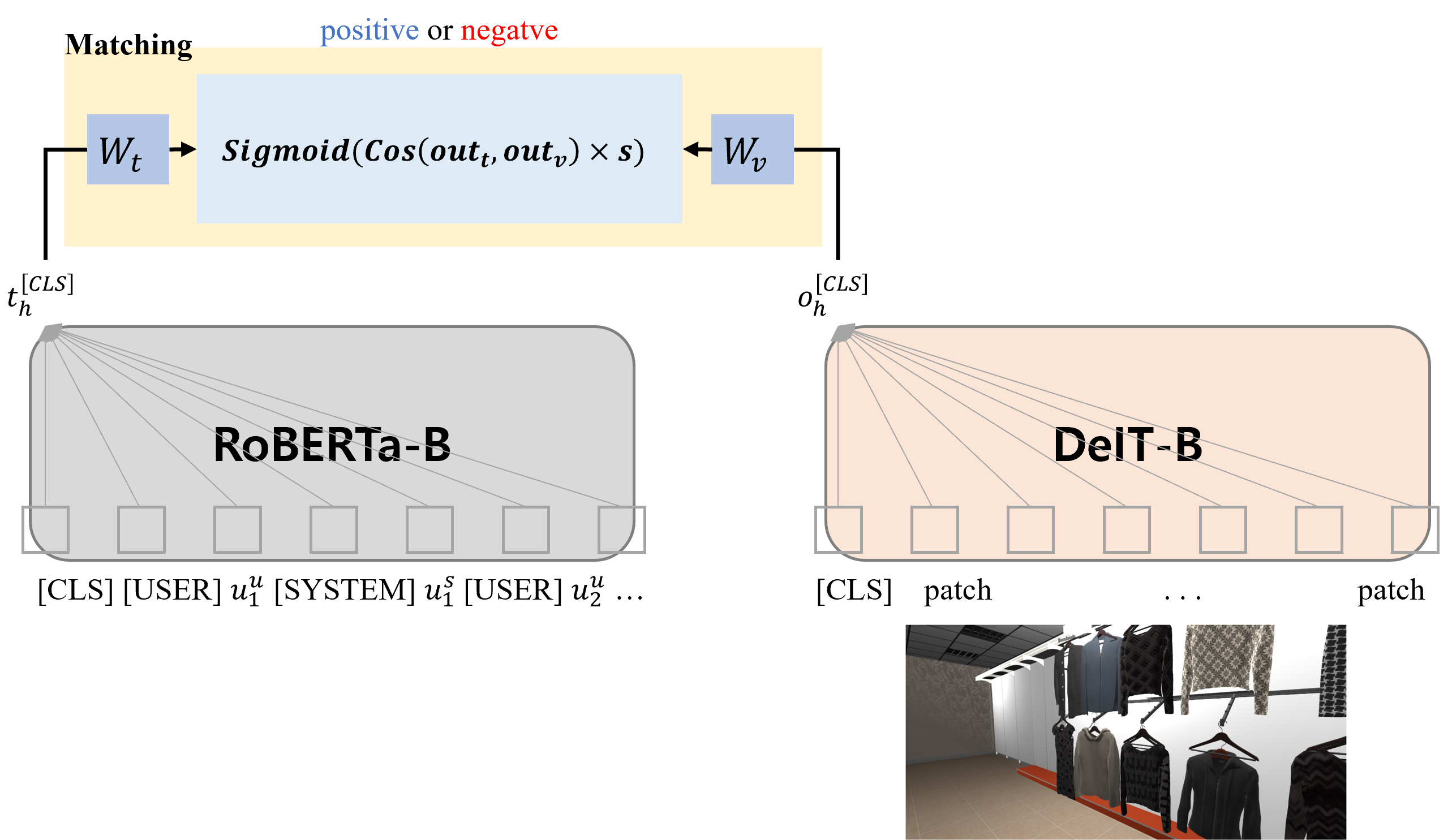}
      \caption{The structure of Background-to-text matching (BTM).}
      \label{fig:sfig2}
    \end{subfigure}
\caption{Multimodal pretraining for mutual understanding of representations between text and image.}
\label{fig:post}
\end{figure*}


In this section, we describe an approach to solving each subtask. First, we pretrain the multimodal model by image-to-text matching to mutually understand the text and image representations before finetuning the subtasks. Then, we train a new model for each subtask based on the pretrained multimodal model. Table~\ref{Tab:notations} shows the notations used in the paper.

\begin{table}[!h]
\centering
\resizebox{1.0\columnwidth}{!}{
\begin{tabular}{cl}
\hline
\textbf{Symbol}  & \textbf{Description}                                                                                                                                        \\ \hline\hline
$i_t$   & input of the text model (RoBERTa or GPT2)                                                                                                              \\ \hline
$i_v$   & input of the image model (DeIT)                                                                                                                        \\ \hline
$t^k_h$ & \begin{tabular}[c]{@{}l@{}}output of text model\\ superscript: the $k$th token or special tokens ({[}CLS{]} or {[}SYSTEM{]})\end{tabular} \\ \hline
$o^k_h$ & \begin{tabular}[c]{@{}l@{}}output of the image model for the object image \\ superscript: the $k$th patch or a special patch ({[}CLS{]})\end{tabular}
\\ \hline
$b^k_h$ & output of the image model for the background image 
\\ \hline
$u^t_k$     & \begin{tabular}[c]{@{}l@{}}$k$th utterance \\ superscript (t): user or system\end{tabular}                                                                                                                             \\ \hline
$\textrm{M}$     & matching function                                                                                                                                  \\ \hline
$W_k$   & matrix of the output layer of subtask \#$k$ ($k$: 1 or 2 or 4)                                                                                                             \\ \hline
\end{tabular}
}
\caption{Notations used in papers}
\label{Tab:notations}
\end{table}

\subsection{Pretraining for Multimodal}
\label{sec:pretraining}
Our unimodal models are based on RoBERTa-Large~\cite{liu2019roberta} and DeIT~\cite{touvron2021training} for text and image, respectively, and the overall structure is shown in Fig.~\ref{fig:post}. If there is no multimodal pretraining for these unimodal models, it is difficult to leverage the pretrained unimodal models because the representations of other modals are not mutually understood. In other words, RoBERTa does not understand the image representation because it is learned only with text data. We show the effect of pretraining through experiments in Section~\ref{sec:experiments}. In addition, since the output of the pretrained multimodal DeIT can be regarded as the output of RoBERTa for the object description, it can be used in various ways. Details are explained in Section~\ref{sec:fine}.

\subsubsection{Matching between object and visual metadata}
\label{sec:ITM}
Our multimodal model is trained to match the cropped object and its description. The description of an object includes visual metadata and non-visual metadata. Non-visual metadata includes keys such as \{customerReview, brand, price, size, materials\}, which can be used in both the training and testing phases, but do not represent the visual of the cropped object. Visual metadata includes keys such as \{assetType, color, pattern, sleeveLength, type\}, which represent the visual of the cropped object, but can only be used in the training phase. As shown in Equation~\ref{eq:post_meta}, the input text ($i_t$) is composed of a concatenation of values corresponding to the key of each visual metadata.

\begin{equation}
\label{eq:post_meta}
i_{t} = Concat([CLS], v(assetType), v(color), ... )
\end{equation}

The input image ($i_v$) is a cropped object and consists of positive and negative samples. Positive samples are objects corresponding to visual metadata, and negative samples are other objects in the same scene.

Inputs corresponding to each modal pass through the text model and image model, respectively, and output a representation vector. $t^{[CLS]}_{h}$ and $o^{[CLS]}_{h}$ are the last hidden vectors of [CLS] (first token (or patch) of $i_t$ and $i_v$) and are representations for text and image, respectively. The multimodal model is trained by calculating the similarity between these two vectors by a matching score function. This trained model is called ITM (image-to-text).

\subsubsection{Matching between background and context}
Background and object images contain different information because background images implicitly contain various objects and their positional relationships. We consider the background to be relevant to all utterances in the dialog. For example, if there are only objects of pants in the background, utterances about coats will not be in the dialog. As shown in  Equation~\ref{eq:post_back}, the text input is the context of the entire dialog, and the image input ($i_v$) is the background image. Negative samples are selected by random sampling among the background images of the training data.

\begin{equation}
\label{eq:post_back}
i_{t} = Concat([CLS], [USER], u^{u}_{1}, [SYSTEM], u^{s}_{1}, ...)
\end{equation}

where [USER] and [SYSTEM] are special tokens prepended to the user and system utterances, and $u^u_k$ and $u^s_k$ are user and system utterances, respectively. The multimodal model for the background is also trained in the same way as ITM described in the section above, and this is called BTM (background-to-text).

\subsubsection{Matching score}
Both ITM and BTM are trained through the same matching score. Matching two vectors is binary classification, and either CE (cross-entropy) or BCE (binary cross-entropy) can be used as a loss. The matching model has an output logit of 2-dimension and 1-dimension in CE and BCE, respectively, and we experimentally observed that BCE loss is better. We first calculate the similarity of two vectors ($t^{[CLS]}_{h}$ and $o^{[CLS]}_{h}$) with the cosine similarity function. Then, by adjusting the similarity in Equation~\ref{eq:matching}, a matching score is calculated with a value between 0 and 1.

\begin{equation}
\begin{split}
\label{eq:matching}
score &= \textrm{M}(t^{[CLS]}_{h}, o^{[CLS]}_{h}) \\
      &= sigmoid(cos(W_t(t^{[CLS]}_{h}), W_v(o^{[CLS]}_{h})) \times s)
\end{split}
\end{equation}

where $W_t$ and $W_v$ are matrices that project $t^{[CLS]}_h$ and $o^{[CLS]}_{h}$ in the same dimension, $s$ is a scale-up constant, 100 is used, and the range of the score is adjusted through sigmoid. 


\subsection{Finetuning for Multimodal}
\label{sec:fine}
\begin{figure*}[!t]
    \centering 
    \includegraphics[width=2.0\columnwidth]{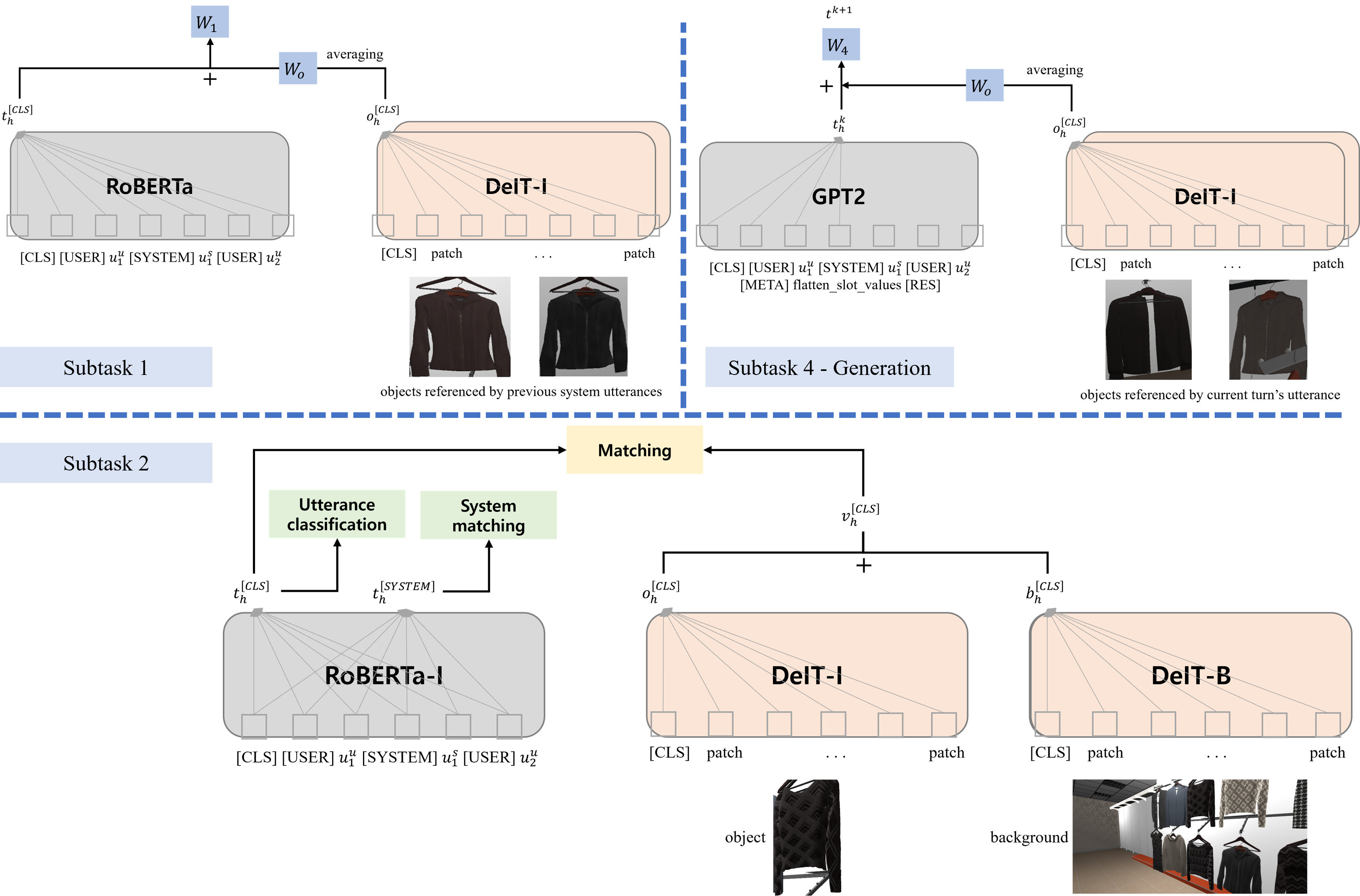}
    \caption{The proposed model architecture. The top left is subtask \#1, the top right is subtask \#4, and the bottom is subtask \#2 architecture. Subtask \#2 includes two multi-task learning.}
    \label{fig:Model}
\end{figure*}

This section introduces finetuning for each subtask using the ITM and BTM in Section~\ref{sec:pretraining}, and the overall structure is shown in Fig.~\ref{fig:Model}.

\subsubsection{Subtask \#1}
Subtask \#1 is a binary classification task for the disambiguation of utterances. Text input is concatenated as in Equation~\ref{eq:post_back} with all utterances before the current turn. The image input is the objects mentioned in the utterances of the previous system, and these objects are accessible for both training and testing. The text representation vector is $t^{[CLS]}_{h}$, the output of RoBERTa, and the image representation vector is the average of $o^{[CLS]}_{h}$, the output of DeIT. The final output follows:

\begin{equation}
\label{eq:sub1}
out_1 = W_1(t^{[CLS]}_{h} + W_o(\sum_{i=1}^{n} o^{[CLS]}_{h}/n)
\end{equation}

where $n$ is the number of objects, $W_o$ is a matrix that projects $o^{[CLS]}_{h}$ in the same dimension as $t^{[CLS]}_{h}$, and $W_1$ is a matrix that projects the final output in two dimensions. In subtask \#1, we use RoBERTa pretrained with text only and DeIT-I trained with ITM. That is, $o^{[CLS]}_{h}$ can be regarded as a representation of the visual metadata of an object. Multimodal of subtask \#1 is trained with CE loss.

\subsubsection{Subtask \#2}
Subtask \#2 has the goal of finding objects referenced by user utterances, similar to ITM. Therefore, we set the backbone of the multimodal model as the pretrained ITM, which is RoBERTa-I and DeIT-I in the subtask \#2 part of Fig.~\ref{fig:Model}. However, unlike ITM, text input in subtask \#2 is the context of a dialog, and the positional relationship between objects in the scene should be considered. Therefore, we use the background image as feature through DeIT-B trained with BTM. The output of DeIT-B is $b^{[CLS]}_{h}$, and the final image representation vector is $o^{[CLS]}_{h}+b^{[CLS]}_{h}$. The match score is calculated as follows:

\begin{equation}
\label{eq:sub2}
score = \textrm{M}(t^{[CLS]}_{h}, o^{[CLS]}_{h}+b^{[CLS]}_{h})
\end{equation}

We introduce two multi-task learning to improve the performance of the model.

1) \textit{Utterance classification}: This task is a binary classification that determines if there is an object to match a given utterance. Label 0 (False) means that the utterance does not refer to any object. The utterances are classified as follows:

\begin{equation}
\label{eq:sub2_utt}
out^{1}_{2} = W^{1}_{2}(t^{[CLS]}_{h})
\end{equation}

2) \textit{System matching}: This multi-task is inspired by the baseline model (GPT2~\cite{radford2019language}) suggested by the challenge organizer. The baseline model matches objects only with context and \textit{object\_ids} corresponding to previous system utterances. The key to this approach is to use previous system utterances that share the objects the current utterance refers to. Therefore, we introduce multi-task learning to find system utterances that share the same objects as the current utterance. System matching is classified as follows.

\begin{equation}
\label{eq:sub2_sys}
out^{2}_{2} = W^{2}_{2}(t^{[SYSTEM]}_{h})
\end{equation}

Two models (+M, -M) are trained depending on whether $mention\_inform$ is used or not, and matching is determined with the logic as shown in Fig.~\ref{fig:pred}. The detailed logic is as follows:

\begin{figure}[!t]
    \centering 
    \includegraphics[width=1.0\columnwidth]{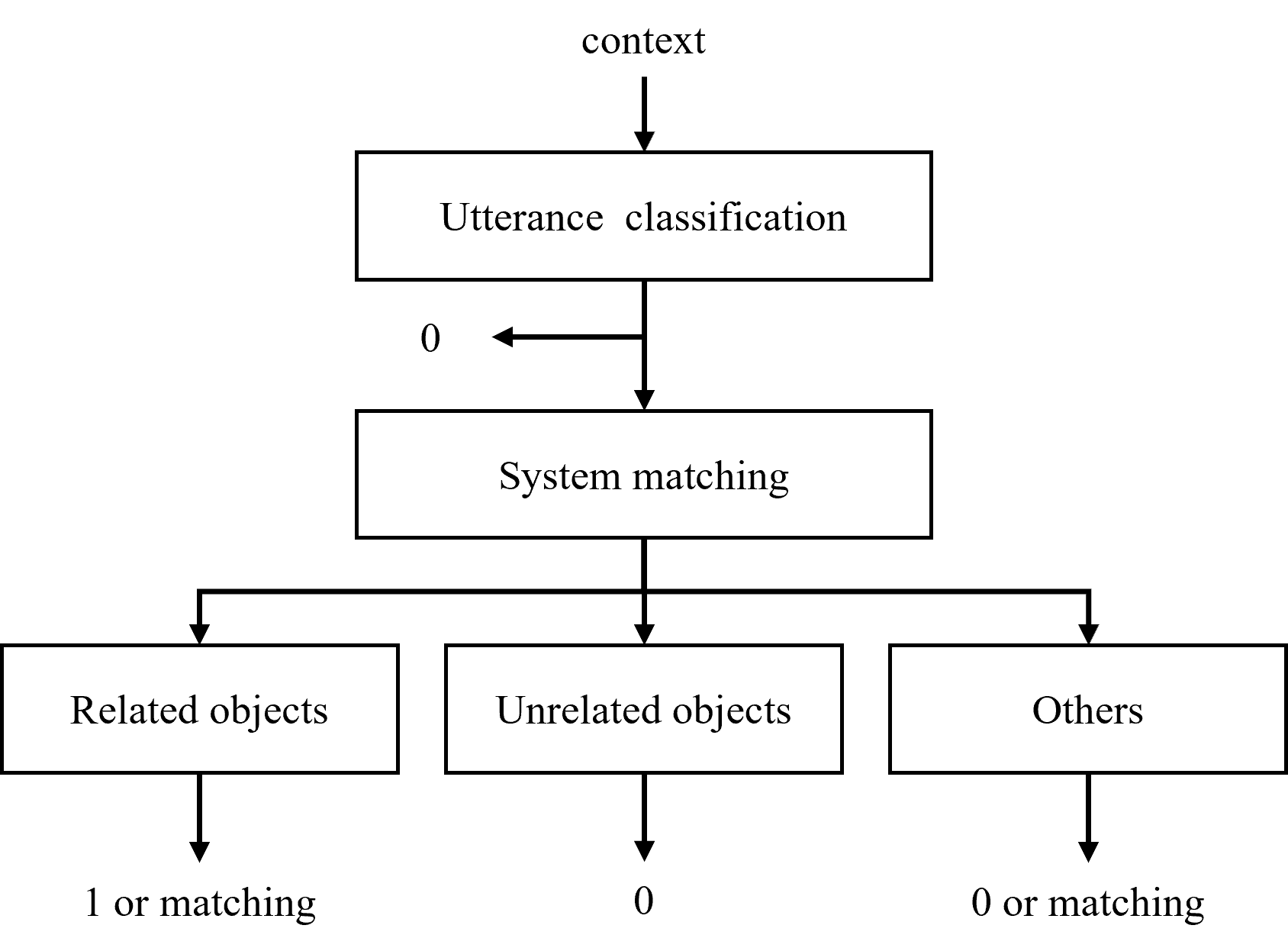}
    \caption{Logic that determines whether the given utterance contains an object in 3 steps through multi-task prediction}
    \label{fig:pred}
\end{figure}

\begin{enumerate}
    \item The model determines whether there are objects to match the current utterance through utterance classification. If utterance classification is not used for training, this step is ignored.
    \item The model finds related objects corresponding to previous system utterances through system matching. If system matching is not used for training, we consider all objects of the previous system utterances to be related objects.
    \item The final prediction depends on the type of models and the category of objects. There are three categories: “related objects”, “unrelated objects”, and “others”. “related objects" are related in a previous system utterance, “unrelated objects" are unrelated in a previous system utterance, and “others" are the rest.
\end{enumerate}

In the statistics of the data, objects that appeared in previous utterances are likely to be matched, and those that did not appear are unlikely to be matched. Accordingly, matching of “related objects" is determined as 1 or a matching score, “unrelated objects" is determined as 0, and “others" is determined as 0 or a matching score. 

If $mention\_inform$ is used for training, the model (Ours+M) encounters more positive objects during training, otherwise the model (Ours-M) encounters more negative objects. Therefore, \textbf{Ours+M} predicts matching of “related objects” through a matching score, and predicts matching of “others” as 0. On the other hand, \textbf{Ours-M} predicts the matching of “related objects” as 1, and predicts the matching of “others” through the matching score. In order to combine the strengths of the two models (\textbf{+M \& -M}), we predict with Ours+M for matching “related objects” and Ours-M for matching “others” to improve performance. \textbf{Ours (only S)} is trained only on the system matching task, not multimodal models, and predicts matching of “related objects" as 1 and the rest as 0.

\subsubsection{Subtask \#4 (Generation)}
Subtask \#4 is a task that generates or retrieves the next turn's system utterance and responds appropriately, and we only experiment with generation. We use GPT2-Large as the text model and DeIT-I as the image model. Unlike other subtasks, \textit{system\_transcript\_annotated} of the current turn can be used. Following the data-to-text generation method mentioned in~\citet{lee-2021-transforming}, we flatten the \textit{slot\_values} in \textit{system\_transcript\_annotated}, and  additionally concatenate “[META] flatten\_slot\_values [RES]" in Equation~\ref{eq:post_back}, where [META] and [RES] are special tokens, and [RES] is a start token for generating utterances. We predict the probability of the next token as in Equation~\ref{eq:sub4} by adding the average of $o^{[CLS]}_{h}$ of the current turn and $t^{k}_{h}$.

\begin{equation}
\label{eq:sub4}
P(t^{k+1}) = \textrm{softmax}(W_{4}(t^{k}_{h}+W_o(\sum_{i=1}^{n} o^{[CLS]}_{h}/n))
\end{equation}

where $W_4$ is a matrix projecting in the dimension of vocabulary size, and $t^{k+1}$ is the (k+1)th token. The utterance is generated using greedy decoding.

\section{Experiments}
\label{sec:experiments}

\subsection{Training Setup}
We use a pretrained unimodal model using the huggingface  library~\footnote{https://huggingface.co/}. The optimizer used for training is AdamW and the learning rate scheduler is $get\_linear\_schedule\_with\_warmup$. We use training epochs from 5 to 10 depending on the subtask. Also, the learning rate is 1e-5 in subtask \#1 and 1e-6 in other subtasks. We use multi-GPU training for two (or four) A100 (or V100) GPUs through apex~\footnote{https://github.com/NVIDIA/apex}.

\subsection{Datasets in the training and testing phases}
\label{sec:data_ours}
\subsubsection{Subtask \#1}
The model is trained only on the data given the disambiguation label in subtask \#1. We observe that utterances such as “Can I get the size and price of that \textit{black top} hanging above the blue pants..." are ambiguous. However, if there is only one “\textit{black top}" in the corresponding image, this utterance will not be ambiguous. In other words, since it is difficult to distinguish such a dataset from text alone, image information is required. However, the objects associated with the current utterance are not accessible data. Instead, we use objects referenced by previous system utterances that are highly relevant to the current utterance and are available in both the training and testing phases.

\subsubsection{Subtask \#2}
Since subtask \#2 has the goal of finding objects that match the user utterances, the objects corresponding to the system utterances are available for testing. However, for more training data, the model is trained to match objects even for system utterances, but at this time, system matching of multi-task learning is not performed, which experimentally shows performance improvement. Also, $mention\_inform$ is a list of all objects mentioned in the dialog, which can be used in the training phase, but cannot be used in the testing phase. We train both models with and without $mention\_inform$ and use them jointly when testing.

\subsubsection{Subtask \#4 (Generation)}
Subtask \#4 has the goal of generating a system utterance for a given context. Similar to subtask \#2, for many training data, the model is also trained to generate user utterances. Also, since subtask \#4 assumes that meta information (\textit{system\_transcript\_annotated}) is known, we use \textit{slot\_values} and \textit{system\_objects} in meta information in both the training and testing phases. We use \textit{slot\_values} because it has properties similar to visual metadata, but not \textit{request\_slots}.

\subsection{Result and Discussion}
Our results are shown in Table~\ref{Tab:result_sub1},~\ref{Tab:result_sub2},~\ref{Tab:result_sub4}. Devtest and teststd are test data during challenge phase 1 and phase 2, respectively. We achieved better performance than baseline models in phase 1 and submitted the best models to phase2. In phase2, we achieved the 3rd best performance in subtask \#1, \#2 and runner-up in the generation part of subtask \#4. 

\subsubsection{Subtask \#1}
In subtask \#1, the models are evaluated with accuracy, and Table~\ref{Tab:result_sub1} shows the results. The baseline is an auto-regressive model based on GPT2. We improved the model performance by combining the bidirectional model RoBERTa and DeIT. However, since DeIT only uses objects that correspond to the previous system utterance, there is room for performance improvement. For example, we expect a higher performance if the model uses objects inferred from subtask \#2 or uses positional information through related objects (or background).

We additionally experimented with multi-task learning that classifies two domains (fashion, furniture) when the model is trained. The accuracy of domain classification achieves about 100\%, but the accuracy of disambiguation classification decreases to 91.2\%.

\begin{table}[!t]
\centering
\resizebox{0.7\columnwidth}{!}{
\begin{tabular}{c|c|c}
\hline
               & \begin{tabular}[c]{@{}c@{}}Accuracy\\      (devtest)\end{tabular} & \begin{tabular}[c]{@{}c@{}}Accuracy\\      (teststd)\end{tabular} \\ \hline\hline
Baseline (GPT2) & 73.9                                                          & 73.5                                                          \\ \hline\hline
Ours        & \textbf{92.28}                                                         & \textbf{93.1}    \\ \hline
\end{tabular}
}
\caption{Accuracy of models in subtask \#1.}
\label{Tab:result_sub1}
\end{table}

\begin{table}[!t]
\centering
\resizebox{0.7\columnwidth}{!}{
\begin{tabular}{c|c|c}
\hline
                   & \begin{tabular}[c]{@{}c@{}}F1 Score\\      (devtest)\end{tabular} & \begin{tabular}[c]{@{}c@{}}F1 Score\\      (teststd)\end{tabular} \\ \hline\hline
Baseline (GPT2)     & 36.6                                                              & 44.1                                                              \\ \hline
Baseline (all   S) & 32.2                                                              & -                                                                 \\ \hline\hline
Ours+M             & 59.5                                                              & 63.4                                                              \\
{\footnotesize -ITM}               & 57.9                                                              & -                                                                 \\
{\footnotesize -BTM}               & 59.4                                                              & -                                                                 \\
{\footnotesize -D}                 & 58.3                                                              & -                                                                 \\
{\footnotesize -U}                 & 56.8                                                              & -                                                                 \\
{\footnotesize -SU}                & 53.5                                                              & -                                                                 \\ \hline\hline
Ours (only S)      & 60.4                                                              & 63                                                                \\ \hline
Ours-M             & 60.7                                                              & 66.7                                                              \\ \hline
Ours (+M \&   -M)  & \textbf{60.8}                                                              & \textbf{68.2}                                                              \\ \hline
\end{tabular}
}
\caption{F1 score of models in subtask \#2. Baseline (all S) is a method that predicts all objects corresponding to the previous system's utterance as true. +M indicates that \textit{mention\_inform} is used, and -M indicates that it is not used. -ITM indicates that our model was trained from random initialization without using a pretrained ITM model. -BTM indicates that our model does not use the pretrained BTM model, but shares the DeIT of the ITM. -D indicates that system utterance and object matching are not used as training data. -U and -S indicate that the model does not use utterance classification and system matching of multi-task learning, respectively.}
\label{Tab:result_sub2}
\end{table}

\subsubsection{Subtask \#2}
In subtask \#2, the models are evaluated with an F1 score, Table~\ref{Tab:result_sub2} shows the results. We introduce a baseline (all S) that predicts all objects referenced by the previous system utterance as true, which performs worse than the baseline (GPT2). Ours (only S) has the disadvantage of not using an image of an object like the baseline (all S), but it is much better than the baseline (all S) and achieves competitive performance with Ours+M. This results support that there are many shared objects between the current utterances and the previous system utterances. 

Table~\ref{Tab:result_sub2} shows the effectiveness of the pretrained multimodal models (ITM, BTM), system utterances as training data (D), and multi-task learning (U, S). We observe that training strategies significantly improve performance in Ours+M. Since ITM is effective in understanding the representations of unimodal models with each other, the performance is improved by +1.6. Multi-task learning improves performance because the proposed model performs high-accuracy filtering on the presence or absence of matching objects and candidates before matching objects. If system utterance is used for training data, the performance is improved by +1.2 due to the effect of data augmentation. In particular, we find that the performance is much improved by comparing only the use of system utterance, excluding other training strategies. Unlike other training strategies, we find that there is no change in performance even when the representation of the background is extracted with DeIT-B instead of DeIT-I. Additionally, in multi-task learning, our model achieves 99.5\% in utterance classification and 93.24 (F1 score) in system matching.

\begin{table}[!t]
\centering
\resizebox{0.7\columnwidth}{!}{
\begin{tabular}{c|c|c}
\hline
               & \begin{tabular}[c]{@{}c@{}}BLEU-4\\      (devtest)\end{tabular} & \begin{tabular}[c]{@{}c@{}}BLEU-4\\      (teststd)\end{tabular} \\ \hline\hline
Baseline (GPT2) & 0.192                                                           & 0.202                                                           \\ \hline
Baseline (MTN)   & 0.217                                                           & 0.211                                                           \\ \hline\hline
Ours           & \textbf{0.285}                                                           & \textbf{0.297}                                                           \\
{\footnotesize -O}             & 0.275                                                           & -                                                               \\
{\footnotesize -M}             & 0.219                                                             & -                                                               \\
{\footnotesize -D}             & 0.281                                                           & -                                                               \\ \hline
\end{tabular}
}
\caption{BLEU-4 score of models in subtask \#4 (Generation). -O indicates that the object's image is not used. -M indicates that the meta information of \textit{slot\_values} is not used. -D indicates that the user's utterance is not used in the training data.}
\label{Tab:result_sub4}
\end{table}

\subsubsection{Subtask \#4 (Generation)}
In subtask \#4, the models are evaluated with a BLEU-4 score, and Table~\ref{Tab:result_sub4} shows the results. We combine \textit{slot\_values} and object representation in GPT2 to achieve better performance than baseline.

We observe that training strategies improve performance through ablation study. In particular, we find that \textit{slot\_values} are an important key for performance, which is inferred because \textit{slot\_values} have similar properties to visual metadata. Additionally, we find out the improvement of +0.01 in combining the representation of the object and +0.004 in learning when user utterance is used.

\section{Conclusion}
We propose a multimodal model based on the unimodal models of RoBERTa and DeIT. Before finetuning the multimodal model to each subtask, we propose a pretraining strategy called ITM and BTM. Then, multimodal interaction is performed for each subtask to improve the model performance. We propose two multitask learning in subtask \#2, which significantly improves the performance. If relation information between objects is combined with the proposed system, further performance improvement is expected. In the challenge, we achieve 3rd best performance in subtask \#1, \#2, and runner-up in subtask \#4-1.

\bibliography{aaai22}


\bigskip

\end{document}